\documentclass[conference]{IEEEtran}
\IEEEoverridecommandlockouts
\usepackage{cite}
\usepackage{amsmath,amssymb,amsfonts}
\usepackage{graphicx}
\usepackage{textcomp}
\usepackage{xcolor}
\def\BibTeX{{\rm B\kern-.05em{\sc i\kern-.025em b}\kern-.08em
    T\kern-.1667em\lower.7ex\hbox{E}\kern-.125emX}}

\usepackage{ulem}

\usepackage{amsmath}
\usepackage{graphicx}
\usepackage{subfig}
\usepackage{flushend}

\usepackage{algorithm}
\usepackage{algpseudocode}
\usepackage{algorithmicx}
\usepackage{float}
\usepackage{geometry}
\usepackage[table]{xcolor}
\usepackage[english]{babel}
\usepackage{microtype}  
\usepackage{float} 
\usepackage{placeins} 
\usepackage{booktabs} 
\usepackage{graphicx} 
\usepackage{multirow}
\usepackage{soul}

\begin{document}

\title{Augmenting Question Answering with A Hybrid RAG Approach}

\author{
Tianyi Yang\IEEEauthorrefmark{1},
Nashrah Haque\IEEEauthorrefmark{1}, Vaishnave Jonnalagadda\IEEEauthorrefmark{1}, Yuya Jeremy Ong\IEEEauthorrefmark{2}$^1$ \\
Zhehui Chen\IEEEauthorrefmark{3}, 
Yanzhao Wu\IEEEauthorrefmark{4}, Lei Yu\IEEEauthorrefmark{5}, 
Divyesh Jadav\IEEEauthorrefmark{6}$^1$ \IEEEcompsocitemizethanks{$^1$ Work done when the authors were with IBM Research.},
Wenqi Wei\IEEEauthorrefmark{1}$^2$ \IEEEcompsocitemizethanks{$^2$ The corresponding author thanks the partial support from Fordham-IBM Research Fellowship and Fordham Faculty Research Grant.}
\\
\IEEEauthorblockA{\IEEEauthorrefmark{1} Fordham University, New York, NY, USA, \{ty10, nhaque14, vj1, wwei23\}@fordham.edu}
\IEEEauthorblockA{\IEEEauthorrefmark{2} Plastic Lab, New York, NY, USA, yuyajeremyong@gmail.com}
\IEEEauthorblockA{\IEEEauthorrefmark{3} Google, Mountain View, California, USA, zchen451@gatech.edu}
\IEEEauthorblockA{\IEEEauthorrefmark{4} Florida International University, Miami, FL, USA, yawu@fiu.edu}
\IEEEauthorblockA{\IEEEauthorrefmark{5} Rensselaer Polytechnic Institute, Troy, NY, USA, yul9@rpi.edu }
\IEEEauthorblockA{\IEEEauthorrefmark{6} Independent Consultant}
}

\maketitle
\thispagestyle{plain}

\begin{abstract}
Retrieval-Augmented Generation (RAG) has emerged as a powerful technique for enhancing the quality of responses in Question-Answering (QA) tasks. However, existing approaches often struggle with retrieving contextually relevant information, leading to incomplete or suboptimal answers. In this paper, we introduce Structured-Semantic RAG (\textbf{SSRAG}), a hybrid architecture that enhances QA quality by integrating query augmentation, agentic routing, and a structured retrieval mechanism combining vector and graph based techniques with context unification. By refining retrieval processes and improving contextual grounding, our approach improves both answer accuracy and informativeness. We conduct extensive evaluations on three popular QA datasets, TruthfulQA, SQuAD and WikiQA, across five Large Language Models (LLMs), demonstrating that our proposed approach consistently improves response quality over standard RAG implementations.
\end{abstract}


\section{Introduction}
Large Language Models (LLMs), such as OpenAI’s GPT series~\cite{achiam2023gpt}, Google’s PaLM~\cite{chowdhery2023palm} and Gemini \cite{team2024gemma}, Meta’s Llama series \cite{Touvron2023LLaMAOA}, and Anthropic’s Claude \cite{TheC3} have significantly contributed towards the advancements in natural language processing (NLP), achieving remarkable performance in applications like question answering, text summarization, machine translation, and sentiment analysis. These advancements have made LLMs indispensable across industries such as finance, healthcare, legal and consulting, where automated decision-making and data retrieval are critical~\cite{eloundou2023gpts,kung2023performance,cui2023chatlaw}.

Despite their capabilities, LLMs often struggle with retrieval and response generation quality, particularly in open-domain question answering. A key limitation is their reliance on parametric memory, which restricts their ability to recall and retrieve factual data dynamically. This issue is further exacerbated by hallucinations~\cite{benkirane2024machine,bommasani2021opportunities}, where LLMs generate responses that appear confident but contain inaccurate or misleading information. Traditional mitigation strategies have sought to enhance factual consistency and reasoning reliability through several complementary methods. For instance, RAG \cite{lewis2020retrieval} augments model prompts with external data sources, thereby grounding responses in verifiable data. Logical reasoning frameworks like Chain of Thought (CoT) \cite{wei2022chain,kojima2022large} decompose multi-step tasks into sequential steps, while Chain of Verification (CoVe) \cite{dhuliawala2024chain} introduces iterative checks against sub-questions to reduce factual errors. Additionally, self-learning or self-refinement mechanisms \cite{ferdinan2024into} allow the model to iteratively evaluate and correct its own responses. Despite these promising techniques, each brings distinct trade-offs. CoT and CoVe often incur high computational overhead during inference \cite{tai2023exploring,dhuliawala2024chain}, and may introduce logical inconsistencies such as self-looping or convergence toward incoherence \cite{stechly2024chain}. RAG, for its part, mitigates hallucinations by anchoring responses in external databases, yet it can falter when retrieval modules fail to capture nuanced semantic or structural context~\cite{martin2024semantic,kim2025towards}, particularly when retrieval relies solely on vector-based methods or simple keyword matching.  

\begin{table*}[t] 
\centering
\caption{SSRAG vs. other RAG systems. \checkmark = strong native support; $\circ$ = partial/indirect; – = absent.}
\begin{tabular}{lcccc}
\toprule
Criterion & \textbf{SSRAG} & RAPTOR & Learned-RAG & Agentic RAG \\
\midrule
Training-free & \textbf{\checkmark} & \checkmark & – & \checkmark \\
Needs labeled/synthetic supervision & – & – & \textbf{\checkmark} & – \\
Real-time freshness (web routing) & \textbf{\checkmark} & – & –/$\circ$ & $\circ$ \\
Explicit factual vs temporal routing & \textbf{\checkmark} & – & – & $\circ$ \\
Graph \& vector merged in one rerank & \textbf{\checkmark} & – & – & – \\
Global dedup to reduce contradictions & \textbf{\checkmark} & – & – & – \\
Precompute-heavy (summary trees / retraining) & – & \textbf{\checkmark} & \textbf{\checkmark} & – \\
\bottomrule
\end{tabular}
\label{table:summary}
\end{table*}

To improve the accuracy and efficiency of LLM-driven question answering, we introduce Structured-Semantic RAG (SSRAG), a hybrid RAG framework that optimizes data retrieval and integration. Unlike standard RAG implementations that rely exclusively on vector-based retrieval, our approach combines vector and graph based retrieval, by unifying them into single context representation, to enhance both semantic precision and structural grounding in retrieved data. Additionally, our framework introduces query understanding and augmentation and agentic query routing modules, which are dynamically refining retrieval queries and directing them to the most relevant data sources. 
Our technical contributions are threefold: 
\begin{itemize}  
\item \textbf{Query Understanding and Augmentation}: We introduce a lightweight prompting-based dynamic query refinement mechanism that expands and clarify user queries.
\item \textbf{Agentic Query Routing}: We propose an rule-guided intelligent query routing mechanism that directs augmented queries to the most relevant data sources from existing fact database or live web search, optimizing efficiency and mitigating retrieval failures.
\item \textbf{Hybrid Retrieval and Context Unification}: We develop a hybrid retrieval strategy that integrates vector and graph-based retrieval techniques to provide richer and more reliable semantic and structural grounding. These hybrid contexts are unified into single context representations that are utilized by an LLM to generate a factually coherent response.  
\end{itemize}

Through extensive evaluations of the TruthfulQA, SQuAD, and WikiQA datasets across five advanced LLMs, our 
proposed framework, SSRAG demonstrates substantial improvements in response accuracy and factual consistency, significantly reducing hallucinations in question-answering tasks. Our findings highlight the potential of hybrid retrieval strategies in augmenting LLM-driven question answering, offering a scalable and adaptable approach to enhancing the trustworthiness of AI-generated responses. 

\section{Related Work}

The mitigation of hallucinations in LLMs has been extensively studied, leading to various strategies to enhance factual accuracy and response reliability. One foundational approach, Chain-of-Thought (CoT) prompting, improves reasoning by decomposing complex problems into sequential steps~\cite{wei2022chain}. CoT has demonstrated strong performance in multi-step reasoning tasks, particularly in mathematics and logic~\cite{kojima2022large}. However, CoT is susceptible to error propagation, where incorrect reasoning at earlier stages can cascade through later steps~\cite{tai2023exploring}. This weakness is magnified in knowledge-intensive tasks, as CoT relies solely on the model’s parametric memory, which may contain outdated or inaccurate information~\cite{lewis2020retrieval}. Furthermore, CoT often struggles with abstract reasoning, particularly when dealing with tasks outside its training data~\cite{kojima2022large}.

To address these issues, Chain-of-Verification (CoVe) introduces an explicit verification phase, where responses are iteratively checked against generated sub-questions and independent verification models~\cite{dhuliawala2024chain}. CoVe reduces factual inconsistencies by breaking complex queries into verifiable units. However, this approach significantly increases computational overhead by requiring multiple inference passes. Other mitigation techniques include self-consistency decoding~\cite{wang2023self}, fact-checking pipelines~\cite{fadeeva2024fact}, and confidence-based filtering~\cite{farquhar2024detecting}. 
Despite these advancement, hallucinations persist, particularly in ambiguous or low-resource data domains~\cite{benkirane2024machine, bommasani2021opportunities}. 

A complementary approach, RAG, enhances reliability by incorporating external data during inference~\cite{lewis2020retrieval}.  Unlike CoT and CoVe, RAG accesses external data sources during inference, allowing retrieval of more up-to-date and verifiable information~\cite{lewis2020retrieval}. Traditional RAG relies on static document retrieval and appends retrieved documents directly to the model’s input context, while adaptive retrieval mechanisms, such as the Layered Query Retrieval framework, dynamically refine sources based on evolving query understanding, improving retrieval accuracy and minimizing exposure to irrelevant content~\cite{huang2024layered}. Recent RAG models have also explored cross-document reasoning, leveraging multiple retrieved sources to verify consistency before response generation~\cite{forer2024inferring}. 
Despite merit, existing RAG systems suffer from semantic drift, where retrieved documents may be loosely relevant but structurally inconsistent, leading to unreliable output~\cite{martin2024semantic}. 

In summary, prior works attempt to optimize in terms of (i) query augmentation, which improves recall but leaves source selection and cross-source consistency unaddressed, (ii) multi-score RAG, which aggregates information from multiple sources but often lacks explicit temporal or factual routing, and (iii) graph-centric RAG, which emphasizes entity-relationship structures but heavily depends on the quality of ingested knowledge graphs or often loses fine-grained details similar to RAPTOR~\cite{RAPTOR2024}. Other RAG systems, such as and Self-RAG require additional model training, which introduces additional overhead~\cite{selfRAG2023, Raft2024}. An independent yet concurrent work~\cite{hybridrag2024} closely related to ours is limited to where the structured knowledge graphs and domain-specific retrieval sources are available, and could struggle to adapt to rapidly evolving data.

\begin{figure*}
    \centering
    \includegraphics[width=0.95\linewidth]{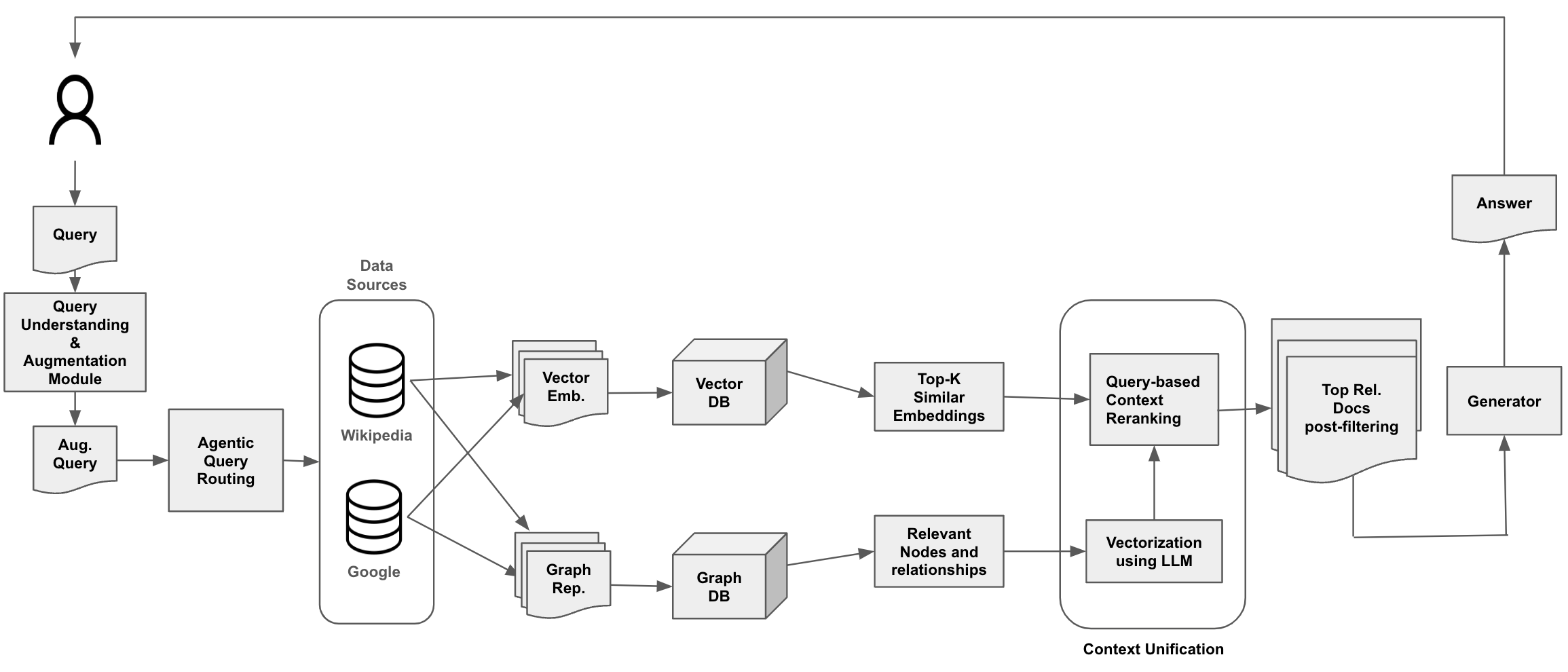}
        \caption{SSRAG Architecture. The user query $Q$ is first augmented and then routed to the most suitable data source. The retrieved context is fused through a hybrid pipeline (graph + vector) before generating the final answer $R$.}
    \label{fig:fig1}
\end{figure*}

In contrast, SSRAG improves traditional RAG models by enhancing retrieval accuracy, factual consistency, and response reliability without the need for model retraining. 
It bridges the gaps of prior approaches by (1) augmenting queries for improved clarity, (2) employing an agentic router that makes explicit, training-free data sourcing decisions, and (3) unifying retrieved content through robust hybrid retrieval and consistency-checking mechanisms. Moreover, its rule-guided routing and single reranking step make it easier to audit and reproduce than fine-tuned agent behaviors, which can drift with changing data distributions. As summarized in \textbf{Table~\ref{table:summary}}, these design choices result in more deterministic, interpretable outputs and establish SSRAG as a flexible and transparent framework for both knowledge-intensive reasoning and general-purpose language modeling.



\section{Methodology}
 This section presents SSRAG, a hybrid RAG framework designed to enhance factual accuracy and reliability in LLM outputs. Unlike conventional RAG implementations, which rely solely on vector-based retrieval, our approach introduces a modular architecture containing three interdependent components: \textit{Query Understanding and Augmentation}, \textit{Agentic Query Routing}, \textit{Hybrid Retrieval Mechanism \& Content Unification}, as depicted in \textbf{Figure~\ref{fig:fig1}}.  Specifically, \textit{Query Understanding and Augmentation} iteratively refines, decomposes, and expands user queries with contextual cues, improving retrieval precision. \textit{Agentic Query Routing} dynamically directs queries to the most appropriate data sources based on semantic and structural context. \textit{Hybrid Retrieval Mechanism \& Content Unification} integrates vector and graph-based retrieval strategies to capture complementary structure and relevance information, and unifies them into single context representation. By optimizing these components in an end-to-end framework, SSRAG systematically improves retrieval effectiveness with more contextually relevant, semantically grounded, and structurally coherent retrieval that mitigates factual inconsistencies in the generated responses. In the following sections, we provide a detailed formulation of each component and its contribution to retrieval efficiency and generation accuracy.


\subsection{Query Understanding and Augmentation}
\label{subsec:query_augment}

\begin{algorithm}[t]
\small
\caption{Query Understanding \& Augmentation}
\label{alg:augment_query}
\begin{algorithmic}[1]
\Function{AugmentQuery}{query}
    \State decompquery $\gets$ DecomposeQuery(query)
    \State entities $\gets$ ExtractKeyEntities(decompquery)
    
    \State intent $\gets$ DetectIntent(decompquery, entities)
    
    \State augQuery $\gets$ EnhanceQuery(query, intent, entities)
    
    \Return augQuery, entities
\EndFunction
\end{algorithmic}
\end{algorithm}


The first stage of our framework, as outlined in \textbf{Algorithm 1}, focuses on refining user queries to enhance retrieval quality through an LLM. The LLM performs \textit{query decomposition}, \textit{key entity extraction}, \textit{intent detection}, and \textit{query augmentation}, generating a semantically enriched query representation, denoted as \( Q_{\text{aug}} \). The process begins with \textit{query decomposition}, where the LLM analyzes the syntactic and semantic structures of the input to identify meaningful components. To extract \textit{key entities}, the LLM is prompted to recognize salient terms that define the query's core intent. Next, the LLM infers \textit{user intent} by interpreting contextual cues, uncovering implicit aspects such as subtopics and retrieval goals. Finally, the LLM reconstructs and enhances the query by integrating extracted entities and inferred intent, the query is augmented, ensuring greater specificity and contextual clarity.

\textbf{Implementation specifics: }
\begin{itemize}
    \item \textbf{Entity extraction:} We prompt an LLM to return a list of named entities (people, organizations, geographic locations, dates, etc.) using a single-turn instruction.
    \item \textbf{Intent cues:} We detect query intent ("who/what/when/where/how") or generate one if one is missing from the original query, as well as if a temporal cue ("today", "currently", "lately", etc) is present or missing, as well as factual cues (proper nouns, historical periods).
    \item \textbf{Augmentation:} Using an LLM, the \textit{EnhanceQuery} step expands and replaces acronyms (for example, "RL" \textrightarrow "reinforcement learning". It also canonicalizes different aliases by mapping through an alias store, and that if multiple surfaces then map to the same canonical but keep the longest surface form (for example, "Churchill/Sir Winston/Sir Winston Churchill" would all be replaced by "Sir Winston Churchill"). Given the extracted entities and intent cues, the LLM would take prompt input: "Given entities=[...], time hints=[...], intent=[...]. Rewrite a single augmented query that: (1) Expands acronyms on first mention in parentheses, (2) Preserves named entities verbatim, (3) Uses $\leq$ 40 tokens."
\end{itemize}

\noindent \textbf{Illustrative example.} Given the query:
\begin{quote}
    \textit{“Albert Einstein's birth place?”}, 
\end{quote}
the system generates an enhanced augmented query by processing the user query as 
\begin{quote}
    \textit{“Where was Albert Einstein born?”}
\end{quote}


\begin{figure*}
    \centering
    \includegraphics[width=0.9\linewidth]{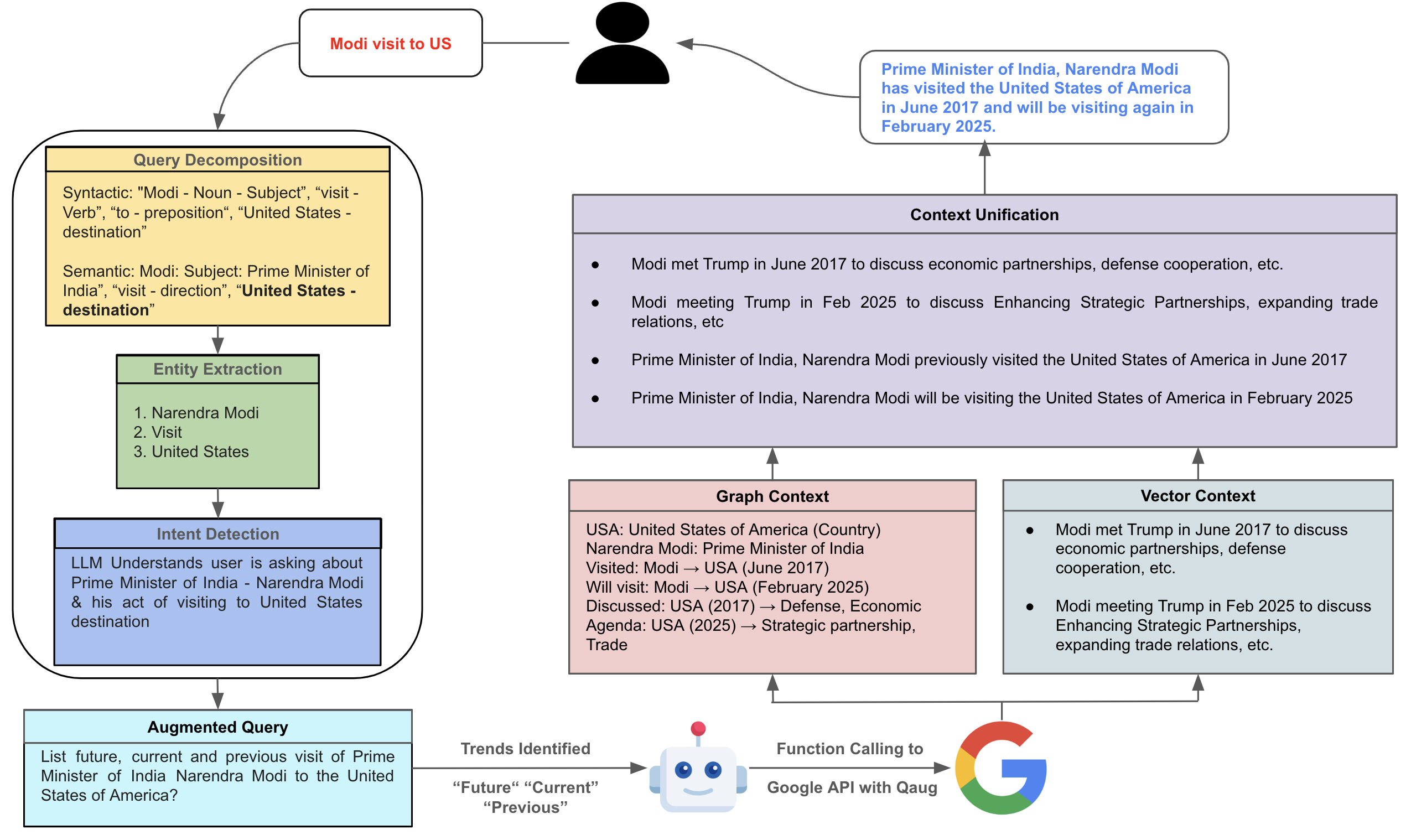}
    \caption{SSRAG: A Comprehensive View of Query Processing, Context Retrieval, and Unification}
    \label{fig:example}
\end{figure*}

\subsection{Agentic Query Routing}  

\label{subsec:query_routing}

\begin{algorithm}[t]
\small
\caption{Agentic Query Routing}
\label{alg:queryrouting}
\begin{algorithmic}[1]
\Function{RouteQuery}{augmentedQuery, entities}
    \If{IsHistoryorFactual(augmentedQuery)}
        \State result $\gets$ QueryWikipediaDB (augmentedQuery)
        \If{result $=$ empty}
            \State route $\gets$ GoogleAPI(augmentedQuery)
        \Else
            \State route $\gets$ WikipediaAPI(entities)
        \EndIf
    \Else
        \State route $\gets$ GoogleAPI(augmentedQuery)
    \EndIf
    
    \Return augmentedQuery, route
\EndFunction
\end{algorithmic}
\end{algorithm}

Once the user query is augmented, the next crucial step is to determine the most appropriate data source for retrieval. This is accomplished through our \textit{agentic query routing} (\textbf{Algorithm~\ref{alg:queryrouting}}) mechanism, which employs a LLM as an intelligent agent to dynamically classify queries and route them to the most relevant external retrieval system. This dynamic classification ensures context-aware routing, reducing the likelihood of retrieving irrelevant or outdated information. Given an augmented query \( Q_{\text{aug}} \), the routing process considers either of the two key factors:
\begin{itemize}
    \item \textbf{Historical/Factual Queries}: Queries containing well-defined entities (e.g., scientific terms, historical figures) are routed to Wikipedia. The system either fetches relevant content from the database or, if data is not available, extracts using the Wikipedia API.
    \item \textbf{Temporal sensitivity}: Time-sensitive queries (e.g., current events, market trends) are directed to real-time sources like Google. Google API is utilized to retrieve the latest indexed web pages relevant to the query. 
\end{itemize}

\textbf{Implementation specifics}, given \textit{Q\_aug}:
\begin{itemize}
    \item If: temporal cues are present (for example, "latest", "in 2023", "today", "this week", "trending"), route to Google Web.
    \item Else: there doesn't exist a temporal cue, then more likelier than not factual cues dominate (for example, named historical figures, scientific concepts, definitions), route to Wikipedia DB corpus.
    \item All routing is prompt-guided and deterministic given \textit{Q\_aug}. The LLM would take input \textit{Q\_aug} and is asked "Decide if this user query is time-sensitive (temporal) or not (factual). Respond with exactly one token: TEMPORAL or FACTUAL." The router output would either point the vector, graph, and hybrid retriever to either web snippets returned by Google API or pre-curated Wikipedia corpus. 
\end{itemize}

\noindent \textbf{Illustrative examples.}
\begin{itemize}
    \item \textbf{“What are the latest breakthroughs in AI?”} \(\rightarrow\) Google, as user query is related to the current events.
    \item \textbf{“Where did Albert Einstein was born?”} \(\rightarrow\)Wikipedia, as user query is related to Historical/Factual events.
\end{itemize}
By integrating Wikipedia Database or API for factual queries and Google API for real-time information retrieval, our approach ensures that queries are matched with the most appropriate data source. This agentic routing process minimizes irrelevant lookups, enhances retrieval accuracy, and improves factual reliability in downstream processing.

\subsection{Hybrid Retrieval \& Context Unification}
\label{subsec:hybrid_retrieval}

To enhance factual consistency and retrieval effectiveness, we propose a \textit{Hybrid Retrieval Mechanism} that unifies \textit{vector} and \textit{graph-based} retrieval. Graph-based retrieval captures entity relationships and structural dependencies, enabling precise reasoning, while vector-based retrieval ensures semantic matching through high-dimensional embeddings. Given a query \( Q \), the system retrieves data from heterogeneous sources using both methods, transforming documents into structured graphs and vector embeddings to preserve both semantic and structural dependencies.

\textbf{Algorithm~\ref{alg:context_unification}} outlines the Hybrid Retrieval \& Context Unification process. Specifically, graph-based context data is first converted into textual representations using an LLM such as GPT-4. These textual representations are then transformed into vector embeddings through an embedding model, effectively vectorizing the nodes and relationships extracted from the graph. Once graph-based vectors are generated, they are combined with existing vector-based embeddings, and the resulting vectors are re-ranked based on their cosine similarity to the user’s query. To ensure the diversity and relevance of the contexts, the top '2k' relevant documents are selected, followed by a deduplication filter where an LLM is employed to eliminate any redundant contexts by evaluating text and cosine similarity scores. Finally, the top k relevant texts are retrieved and used as a context for generating the final response. 

\noindent \textbf{Illustrative example.} Given a query \textit{“How does reinforcement learning apply to robotics?”}, the retrieval operates in parallel:
\begin{itemize}
    \item \textbf{Graph-based retrieval}: Extracts structured relationships between entities such as \textit{Reinforcement Learning}, \textit{Robotics}, and \textit{Policy Optimization}.
    \item \textbf{Vector-based retrieval}: Computes semantic embeddings to identify relevant documents.
    \item \textbf{Joint rerank}: Concatenate vector candidates and graph-to-text candidates, score each by cosine similarity. Select \textit{Top-2k} for redundancy control. Then do \textit{TextDedup} to remove near-duplicates, specifically exact-match removal and high cosine tie-break. Finally, take \textit{Top-k} as the unified evidence set.
\end{itemize}


%
%
%

\begin{algorithm}[t]
\small
\caption{Hybrid Retrieval \& Context Unification}
\label{alg:context_unification}
\begin{algorithmic}[1]
\Function{ContextUnification}{topkEmbeds, relevantNodes, relations, embedModel, QueryVector (Qv)}
    \State CombVectors (Cv) $\gets$ topkEmbeds
    
    \For{each {\underline{node}} in relevantNodes}
        \State nodeRel $\gets$ GetRelation(node, relations)
        \State textRep $\gets$ LLM\_Convert(node, nodeRel)
        \State  vector $\gets$ embedModel.vectorize(textRep)
        \State Cv.add(vector)
    \EndFor

    \State rel\_vectors, SimScores $\gets$ GetTop2k(Qv, Cv)
    \State retrievedtexts $\gets$ GetTexts(rel\_vectors)
    \State filteredtexts $\gets$ TextDedup(retrievedtexts, SimScores)
    
    \State \Return filteredtexts
\EndFunction
\end{algorithmic}
\end{algorithm}

\textbf{An Illustrative Example of SSRAG Framework.} \textbf{Figure \ref{fig:example}} Illustrates an example of our proposed framework. The user query, “Modi visit to US,” is inherently ambiguous as it does not explicitly indicate a specific timeline—whether referring to a past, present, or future event. Unlike ChatGPT, which exhibits recency bias by assuming queries are related to current events and responding accordingly, our approach makes no such assumptions. Instead, it systematically considers all possible timelines to generate a more comprehensive and contextually enriched response. Furthermore, the context unification phase leverages a diverse set of structured and unstructured data sources, ensuring a high-quality and well-informed response to the user.

\subsection{Implementation Details}
\label{subsec:technical_implementation}

The implementation of SSRAG follows a modular design that integrates vector and graph-based retrieval into a scalable architecture. This section details the implementation of each component.

\textbf{Vector-Based Retrieval} 
enables efficient similarity search by encoding documents into high-dimensional embeddings. Given a query, the system performs a nearest-neighbor search over a precomputed Facebook AI Similarity Search (FAISS) \cite{douze2024faisslibrary} vector index to retrieve semantically relevant documents\cite{johnson2019billion}. Specifically, the query will be embedded using OpenAI’s \textit{text-embedding-\allowbreak3-small} model and matched against stored document embeddings. The process consists of the following steps:




\begin{itemize}
    \item \textbf{Data Processing}: Text is extracted and preprocessed using LangChain’s text-splitting.
    \item \textbf{Embedding Generation}: Document chunks are transformed into dense vectors using an OpenAI embedding model.
    \item \textbf{Vector Indexing}: FAISS stores and indexes embeddings for optimized retrieval.
    \item \textbf{Similarity Search}: Queries are encoded and matched against stored vectors.
    \item \textbf{LLM Integration}: Retrieved documents provide external context for response generation.
\end{itemize}


\textbf{Graph-Based Retrieval}
 extracts structured knowledge by representing information as an entity-relationship graph, enabling relational reasoning for complex queries. Instead of retrieving semantically similar text, the system performs:
\begin{itemize}
    \item \textbf{Entity Extraction}: Identifies key concepts (e.g., \textit{Reinforcement Learning, Robotics, Policy Optimization}).
    \item \textbf{Graph Construction}: Constructs an entity-relationship graph in Neo4j.
    \item \textbf{Query Execution}: Uses Cypher-based queries to retrieve relevant subgraphs.
    \item \textbf{Context Augmentation}: Passes structured graph output to the LLM.
\end{itemize}



\begin{table*}[h!]
\centering
\caption{Baseline RAG vs. Graph RAG vs. SSRAG on Truthful QA Dataset}
\label{tab:model-performance-truthfulQA}
\scalebox{0.9}{
\begin{tabular}{ccccccccccc}
\hline
    Model                        & Metric              & \begin{tabular}[c]{@{}c@{}}Baseline \\ RAG\end{tabular} & \begin{tabular}[c]{@{}c@{}}Graph \\ RAG\end{tabular} & \begin{tabular}[c]{@{}c@{}}SSRAG\end{tabular} & CoT     & \begin{tabular}[c]{@{}c@{}}CoT with \\ Graph RAG\end{tabular} & \begin{tabular}[c]{@{}c@{}}CoT with \\ SSRAG\end{tabular} & CoVe    & \begin{tabular}[c]{@{}c@{}}CoVe \\ with Graph\end{tabular} & \begin{tabular}[c]{@{}c@{}}CoVe with \\ SSRAG\end{tabular} \\ \hline
\multirow{4}{*}{GPT-4}       & \% true (GPT-judge) & 57\%                                                    & 67\%                                                 & \textbf{87\%}                                              & 57\%    & 63\%                                                          & \textbf{81\%}                                                       & 58\%    & 64\%                                                       & \textbf{83\%}             \\
                             & ROUGE-1               & 43.2\%                                                   & 58\%                                                 & \textbf{82.3\%}                                             & 47\%    & 54.7\%                                                         & \textbf{83.4\%}                                                      & 48\%    & 57.7\%                                                      & \textbf{77\%}             \\
                             & BLEU-1                & 47\%                                                    & 65\%                                                 & \textbf{86.4\%}                                            & 45.4\% & 62.7\%                                                       & \textbf{82\%}                                                       & 41.1\% & 57.9\%                                                    & \textbf{83\%}             \\
                             & Self CheckGPT       & 0.51                                                    & 0.4                                                  & \textbf{0.18}                                              & 0.47    & 0.37                                                          & \textbf{0.16}                                                       & 0.47    & 0.42                                                       & \textbf{0.15}             \\ \hline
\multirow{4}{*}{LLaMA 2}     & \% true (GPT-judge) & 52\%                                                    & 65\%                                                 & \textbf{83\%}                                              & 52\%    & 58\%                                                          & \textbf{88\%}                                                       & 53\%    & 59\%                                                       & \textbf{87\%}             \\
                             & ROUGE-1               & 52.1\%                                                   & 59.8\%                                                & \textbf{75.2\%}                                             & 49.2\%   & 53.3\%                                                         & \textbf{81.2\%}                                                      & 51\%    & 58\%                                                       & \textbf{79.2\%}            \\
                             & BLEU-1                & 52\%                                                    & 68\%                                                 & \textbf{82.4\%}                                           & 47.3\% & 63.1\%                                                       & \textbf{87.1\%}                                                    & 45.1\% & 61.1\%                                                    & \textbf{85\%}             \\
                             & Self CheckGPT       & 0.52                                                    & 0.38                                                 & \textbf{0.23}                                              & 0.51    & 0.41                                                          & \textbf{0.18}                                                       & 0.54    & 0.37                                                       & \textbf{0.2}              \\ \hline
\multirow{4}{*}{TinyLLaMA}   & \% true (GPT-judge) & 47\%                                                    & 57\%                                                 & \textbf{75\%}                                              & 46\%    & 52\%                                                          & \textbf{72\%}                                                       & 47\%    & 53\%                                                       & \textbf{72\%}             \\
                             & ROUGE-1               & 32.1\%                                                   & 47\%                                                 & \textbf{81.1\%}                                             & 51.2\%   & 58.7\%                                                         & \textbf{79\%}                                                       & 56\%    & 59\%                                                       & \textbf{73\%}             \\
                             & BLEU-1                & 41\%                                                    & 51\%                                                 & \textbf{91\%}                                           & 53.5\% & 67.8\%                                                       & \textbf{79\%}                                                    & 52.2\% & 55\%                                                       & \textbf{81\%}          \\
                             & Self CheckGPT       & 0.54                                                    & 0.47                                                 & \textbf{0.27}                                              & 0.56    & 0.44                                                          & \textbf{0.29}                                                       & 0.49    & 0.41                                                       & \textbf{0.29}             \\ \hline
\multirow{4}{*}{IBM Granite} & \% true (GPT-judge) & 52\%                                                    & 61\%                                                 & \textbf{79\%}                                              & 47\%    & 53\%                                                          & \textbf{69\%}                                                       & 48\%    & 54\%                                                       & \textbf{72\%}             \\
                             & ROUGE-1               & 51.4\%                                                   & 56\%                                                 & \textbf{79.8\%}                                             & 56.7\%   & 57.2\%                                                         & \textbf{71\%}                                                       & 55\%    & 57\%                                                       & \textbf{76\%}             \\
                             & BLEU-1                & 53\%                                                    & 61\%                                                 & \textbf{89.1\%}                                           & 51.1\% & 64.8\%                                                       & \textbf{81\%}                                                    & 51.1\% & 57.2\%                                                    & \textbf{83.2\%}          \\
                             & Self CheckGPT       & 0.45                                                    & 0.42                                                 & \textbf{0.19}                                              & 0.46    & 0.33                                                          & \textbf{0.23}                                                       & 0.57    & 0.38                                                       & \textbf{0.21}             \\ \hline
\multirow{4}{*}{Gemini 1.5}  & \% true (GPT-judge) & 56\%                                                    & 67\%                                                 & \textbf{82\%}                                              & 48\%    & 54\%                                                          & \textbf{78\%}                                                       & 49\%    & 55\%                                                       & \textbf{79\%}             \\
                             & ROUGE-1               & 50\%                                                     & 59\%                                                 & \textbf{85.2\%}                                             & 50\%     & 62.1\%                                                         & \textbf{76\%}                                                       & 49.8\%   & 52\%                                                       & \textbf{77\%}             \\
                             & BLEU-1                & 55\%                                                    & 58\%                                                 & \textbf{91.6\%}                                           & 52.67\% & 65.7\%                                                       & \textbf{87\%}                                                    & 52.67\% & 57.4\%                                                    & \textbf{87.2\%}          \\
                             & Self CheckGPT       & 0.49                                                    & 0.333                                                & \textbf{0.29}                                              & 0.48    & \textbf{0.31}                                                 & 0.32                                                                & 0.48    & 0.32                                                       & \textbf{0.3}              \\ \hline
\end{tabular}}
\end{table*}


\textbf{SSRAG-A hybrid approach.}   
The final stage combines both retrieval strategies into a unified framework, and performs,
\begin{itemize}
    \item \textbf{Parallel Processing}: Queries are processed through both retrieval pipelines.
    \item \textbf{Context Unification}: Retrieved data is merged into a unified representation.
    \item \textbf{LLM-Based Filtering}: Removes redundant context before response generation.
    \item \textbf{Final Response}: The hybrid contexts are passed to the LLM for generation.
\end{itemize}
Our approach leverages both structured and unstructured data, improving retrieval precision and response quality.

\section{Experiments \& Results}
\label{sec:experiments}
To evaluate the effectiveness of the proposed SSRAG framework,
we conduct experiments on three benchmark Q\&A datasets across five large language models. 
This section describes the experimental setup, presents the main results, and provides qualitative insights through case studies.

\subsection{Data, Models \& Evaluation Metrics}
\label{sec:setup}

\noindent \textbf{Datasets.} Our experiments leverage three benchmark datasets that collectively assess truthfulness, fact verification, and contextual reasoning in open-domain settings: {TruthfulQA} as released in 2022~\cite{lin-etal-2022-truthfulqa}, {SQuAD (Stanford Question Answering Dataset)} v2.0~\cite{rajpurkar2016squad} and WikiQA  (2015)~\cite{yang2015wikiqa}. For each dataset we sample 900 questions uniformly at random (seed = 42) to evaluate the LLM's generated responses. Context from Wikipedia snapshot as of February 14, 2025 was extracted and stored in both Vector and Graph databases.

\noindent \textbf{LLMs and Reasoning Frameworks.} 
We assess the performance of SSRAG using five LLMs alongside two  reasoning-based frameworks Chain-of-Thought (CoT)~\cite{wei2022chain} and Chain-of-Verification (CoVe)~\cite{dhuliawala2024chain}. Specifically, we evaluate OpenAI’s GPT-4~\cite{achiam2023gpt}, Meta’s Llama 2~\cite{touvron2023llama2}, TinyLlama~\cite{min2023tinyllama}, IBM Granite Model~\cite{IBMGranite2023}, and Google Gemini~\cite{googlegemini2023}.

\noindent \textbf{Evaluation Metrics.} We consider a set of metrics covering linguistic quality, 
factual consistency, and retrieval effectiveness of LLM generation. \textit{BLEU-1}~\cite{Papineni2002Bleu} 
measures the precision of matching n-grams (typically 1-4) between system outputs and reference texts, 
where higher scores [0-1] indicate better alignment with reference words and more relevant content 
in the generated output. \textit{ROUGE-1}~\cite{lin-2004-rouge} evaluates n-gram recall, 
with higher scores [0-1] reflecting better capture of contextually relevant words in the output. 
These two metrics assess the \textit{retrieval segment} of the RAG architecture. By comparison, 
\textit{\%True GPT Judge} measures factual accuracy, with higher scores [0-100\%] showing better 
alignment with ground truth. \textit{SelfCheckGPT}~\cite{manakul2023selfcheckgpt} 
detects hallucinations [0-1], where a score closer to 1 indicates more inconsistencies in the model’s output. 
These metrics evaluate the \textit{generation segment} of RAG. In addition, we consider 
\textit{RAGAS (Retrieval-Augmented Generation Assessment Score)}~\cite{es2024ragas}, 
a multi-faceted metric designed for RAG systems. RAGAS combines retrieval precision, 
contextual alignment, and factual correctness.

\begin{table*}[h!]
\centering
\caption{Baseline RAG vs. Graph RAG vs. SSRAG on SQuAD Dataset}
\label{tab:model-performance}
\scalebox{0.9}{
\begin{tabular}{ccccccccccc}
\hline
Model                        & Metric              & \multicolumn{1}{c}{\begin{tabular}[c]{@{}c@{}}Baseline \\ RAG\end{tabular}} & \multicolumn{1}{c}{\begin{tabular}[c]{@{}c@{}}Graph \\ RAG\end{tabular}} & \multicolumn{1}{c}{\begin{tabular}[c]{@{}c@{}}SSRAG\end{tabular}} & \multicolumn{1}{c}{CoT} & \multicolumn{1}{c}{\begin{tabular}[c]{@{}c@{}}CoT with \\ Graph RAG\end{tabular}} & \multicolumn{1}{c}{\begin{tabular}[c]{@{}c@{}}CoT with \\ SSRAG\end{tabular}} & \multicolumn{1}{c}{CoVe} & \multicolumn{1}{c}{\begin{tabular}[c]{@{}c@{}}CoVe \\ with Graph\end{tabular}} & \multicolumn{1}{c}{\begin{tabular}[c]{@{}c@{}}CoVe with \\ SSRAG\end{tabular}} \\ \hline
\multirow{4}{*}{GPT-4}       & \% true (GPT-judge) & 54\%                                                                        & 66\%                                                                     & \textbf{84\%}                                                                  & 56\%                    & 62\%                                                                              & \textbf{83\%}                                                                           & 58\%                     & 63\%                                                                           & \textbf{84\%}                                 \\
                             & ROUGE-1               & 43.2\%                                                                       & 58\%                                                                     & \textbf{82.3\%}                                                                 & 47\%                    & 54.7\%                                                                             & \textbf{83.4\%}                                                                          & 48\%                     & 57.7\%                                                                          & \textbf{77\%}                                 \\
                             & BLEU-1                & 47\%                                                                        & 65\%                                                                     & \textbf{86\%}                                                                  & 45\%                    & 63\%                                                                              & \textbf{82\%}                                                                           & 41\%                     & 58\%                                                                           & \textbf{83\%}                                 \\
                             & Self CheckGPT       & 0.52                                                                        & 0.37                                                                     & \textbf{0.15}                                                                  & 0.43                    & 0.33                                                                              & \textbf{0.13}                                                                           & 0.47                     & 0.39                                                                           & \textbf{0.13}                                 \\ \hline
\multirow{4}{*}{LLaMA 2}     & \% true (GPT-judge) & 53\%                                                                        & 62\%                                                                     & \textbf{82\%}                                                               & 51\%                 & 57\%                                                                           & \textbf{89\%}                                                                        & 52\%                  & 58\%                                                                        & \textbf{89\%}                              \\
                             & ROUGE-1               & 49.9\%                                                                       & 54\%                                                                     & \textbf{82.3\%}                                                                 & 47.8\%                   & 58.2\%                                                                             & \textbf{79\%}                                                                           & 52.3\%                    & 56\%                                                                           & \textbf{81.2\%}                                \\
                             & BLEU-1               & 51\%                                                                        & 63\%                                                                     & \textbf{90\%}                                                                  & 51\%                    & 66\%                                                                              & \textbf{80\%}                                                                           & 50\%                     & 60\%                                                                           & \textbf{86\%}                                 \\
                             & Self CheckGPT       & 0.53                                                                        & 0.32                                                                     & \textbf{0.21}                                                                  & 0.49                    & 0.39                                                                              & \textbf{0.2}                                                                            & 0.52                     & 0.38                                                                           & \textbf{0.198}                                \\ \hline
\multirow{4}{*}{TinyLLaMA}   & \% true (GPT-judge) & 49\%                                                                        & 52\%                                                                     & \textbf{74\%}                                                               & 43\%                 & 49\%                                                                           & \textbf{74\%}                                                                        & 44\%                  & 50\%                                                                        & \textbf{69\%}                              \\
                             & ROUGE-1               & 31.5\%                                                                       & 49\%                                                                     & \textbf{85.2\%}                                                                 & 49.1\%                   & 59.9\%                                                                             & \textbf{77\%}                                                                           & 51\%                     & 57\%                                                                           & \textbf{78\%}                                 \\
                             & BLEU-1                & 47\%                                                                        & 58\%                                                                     & \textbf{96\%}                                                                  & 54\%                    & 66\%                                                                              & \textbf{84\%}                                                                           & 54\%                     & 58\%                                                                           & \textbf{87\%}                                 \\
                             & Self CheckGPT       & 0.47                                                                        & 0.41                                                                     & \textbf{0.29}                                                                  & 0.52                    & 0.41                                                                              & \textbf{0.29}                                                                           & 0.59                     & 0.39                                                                           & \textbf{0.3}                                  \\ \hline
\multirow{4}{*}{IBM Granite} & \% true (GPT-judge) & 51\%                                                                        & 63\%                                                                     & \textbf{81\%}                                                               & 50\%                 & 56\%                                                                           & \textbf{76\%}                                                                        & 51\%                  & 57\%                                                                        & \textbf{77\%}                              \\
                             & ROUGE-1               & 52\%                                                                        & 57\%                                                                     & \textbf{81.2\%}                                                                 & 54.3\%                   & 59.9\%                                                                             & \textbf{76.5\%}                                                                          & 53\%                     & 56\%                                                                           & \textbf{74\%}                                 \\
                             & BLEU-1                & 54\%                                                                        & 59\%                                                                     & \textbf{94\%}                                                                  & 50\%                    & 69\%                                                                              & \textbf{85\%}                                                                           & 50\%                     & 59\%                                                                           & \textbf{87\%}                                 \\
                             & Self CheckGPT       & 0.42                                                                        & 0.39                                                                     & \textbf{0.21}                                                                  & 0.43                    & 0.32                                                                              & \textbf{0.23}                                                                           & 0.511                    & 0.37                                                                           & \textbf{0.22}                                 \\ \hline
\multirow{4}{*}{Gemini 1.5}  & \% true (GPT-judge) & 53\%                                                                        & 65\%                                                                     & \textbf{83\%}                                                               & 49\%                 & 55\%                                                                           & \textbf{81\%}                                                                        & 50\%                  & 56\%                                                                        & \textbf{78\%}                              \\
                             & ROUGE-1               & 49.7\%                                                                       & 56\%                                                                     & \textbf{78.9\%}                                                                 & 49.8\%                   & 59.8\%                                                                             & \textbf{71\%}                                                                           & 47.7\%                    & 54\%                                                                           & \textbf{72\%}                                 \\
                             & BLEU-1                & 56\%                                                                        & 57\%                                                                     & \textbf{93\%}                                                                  & 51\%                    & 69\%                                                                              & \textbf{87\%}                                                                           & 51\%                     & 60\%                                                                           & \textbf{86\%}                                 \\
                             & Self CheckGPT       & 0.41                                                                        & 0.43                                                                     & \textbf{0.27}                                                                  & 0.46                    & \textbf{0.3}                                                                      & \textbf{0.28}                                                                           & 0.54                     & 0.31                                                                           & \textbf{0.28}                                 \\ \hline
\end{tabular}}
\end{table*}

\subsection{Main Results}
\label{sec:main_results}

Our proposed SSRAG framework integrates \textit{graph-based} and \textit{vector-based} retrieval mechanisms to enhance contextual relevance, reasoning clarity, and factual consistency in LLMs. Our results demonstrate 
that SSRAG consistently outperforms conventional RAG models and standalone retrieval systems on three popular QA datasets, TruthfulQA, SQuAD and WikiQA, across
five LLMs. By leveraging structured relationships and semantic representations, our approach achieves significant improvements in faithfulness, answer relevancy, and context precision, reinforcing its potential as a robust solution for mitigating hallucinations and enhancing factual grounding.

\noindent \textbf{Improvements Over Existing RAG Frameworks.} 
We first compare our SSRAG approach against the Baseline RAG and Graph RAG methods, focusing on GPT-4’s performance across TruthfulQA (\textbf{Table~\ref{tab:model-performance-truthfulQA}}) and SQuAD (\textbf{Table~\ref{tab:model-performance}}). On TruthfulQA, SSRAG improves \%\,True (GPT-judge) from 57\% (Baseline) and 67\% (Graph) to 87\%, marking a 30\% boost over the Baseline and 20\% over Graph. We observe similar gains in ROUGE-1 (0.432 $\rightarrow$ 0.58 $\rightarrow$ 0.823) and BLEU-1 (47\% $\rightarrow$ 65\% $\rightarrow$ 86.4\%). Additionally, SelfCheckGPT decreases to 0.18—significantly lower than 0.51 (Baseline) and 0.4 (Graph), indicating fewer hallucinations.

On the SQuAD dataset, SSRAG likewise outperforms both Baseline vector RAG and Graph RAG. Specifically, \%\,True (GPT-judge) improves from 54\% to 66\% to 84\%, while ROUGE-1 jumps from 0.432 and 0.58 to 0.823. BLEU-1 follows a similar trend (47\% $\rightarrow$ 65\% $\rightarrow$ 86\%), and SelfCheckGPT is minimized at 0.15, substantially below the 0.52 and 0.37 observed with baseline and graph, respectively. These consistent improvements highlight that SSRAG yields both stronger factuality and fewer hallucinations.

Further breakdowns across different models reinforce these trends. For instance, TinyLLaMA and IBM Granite show substantial factuality gains when using SSRAG, with IBM Granite achieving an 83.2\% BLEU-1 score on TruthfulQA, a significant leap from 53\% (Baseline) and 61\% (Graph). Similarly, TinyLLaMA sees its BLEU-1 score rise from 41.5\% (Baseline) and 51.9\% (Graph) to 91.1\%, demonstrating that SSRAG enhances both small and large models alike.

\noindent \textbf{Enhancing Reasoning with Chain of Thought (CoT) and Chain of Verification (CoVe).} 
Next, we evaluate how SSRAG interacts with well-known reasoning strategies CoT and CoVe, again focusing on GPT-4 for the TruthfulQA and SQuAD tasks. On TruthfulQA, CoT with Baseline RAG achieves 57\%, \%true (GPT-judge) while CoT with Graph RAG improves to 63\%. In contrast, CoT with SSRAG reaches 81\%, surpassing Baseline and Graph by 24 and 18 points, respectively. This pattern extends across LLaMA 2, TinyLLaMA, and IBM Granite, all of which see at least a 15-20\% boost when CoT is applied alongside SSRAG. A similar trend appears under CoVe (58\% $\rightarrow$ 64\% $\rightarrow$ 83\%), demonstrating that SSRAG meaningfully boosts factual accuracy.

We see parallel gains on SQuAD. CoT with SSRAG attains 83\% \%\,True (GPT-judge), notably higher than the 56\% and 62\% for CoT with Baseline and Graph RAG. Likewise, CoVe jumps from 58\% and 63\% to 84\%, a strong indication that combining SSRAG with reasoning frameworks reduces inaccuracies across multiple data sets.

\begin{table*}[t]
  \centering
  \caption{RAGAS Evaluation of SSRAG Framework on WikiQA Dataset}
  \label{tab:model-evaluation-metrics}
  \begin{tabular}{lcccc}
    \toprule
    Model            & Faithfulness & Answer Relevancy & Context Relevancy & Context Precision \\
    \midrule
    LLAMA2           & 92\%        & 89\%             & 91\%              & 88\%              \\
    GPT-4           & 88\%        & 87\%             & 86\%              & 85\%              \\
    IBM Granite 3.0  & 88\%        & 87\%             & 86\%              & 85\%              \\
    TinyLLAMA        & 80\%        & 78\%             & 79\%              & 77\%              \\
    Gemini 1.5 pro   & 88\%        & 87\%             & 86\%              & 85\%              \\
    \bottomrule
  \end{tabular}
\end{table*}

\begin{table*}[t]
  \centering
  \caption{Sensitivity on the number of retrieved documents on WikiQA Dataset}
  \label{tab:hyperparameter-evaluation}
  \begin{tabular}{lcccc}
    \toprule
    Hyperparameter   & Faithfulness & Answer Relevancy & Context Relevancy & Context Precision \\
    \midrule
    N = 5            & 75\%        & 77\%             & 79\%              & 75\%              \\
    N = 10           & 79\%        & 81\%             & 80\%              & 83\%              \\
    N = 15           & 82\%        & 85\%             & 90\%              & 91\%              \\
    N = 20           & 91\%        & 89\%             & 92\%              & 92\%              \\
    \bottomrule
  \end{tabular}
\end{table*}

Moreover, the consistent reduction in SelfCheckGPT scores across CoT and CoVe setups further validates that SSRAG not only improves retrieval accuracy but also decreases hallucinated completions, resulting in more factually sound and contextually relevant outputs.

\noindent \textbf{Consistent Enhancement Across Different LLMs and Multiple Datasets.}
To thoroughly investigate whether SSRAG can benefit diverse model architectures and use cases, we evaluate it on five LLMs that span a range of parameter sizes, training objectives, and deployment settings: GPT-4, LLaMA 2, TinyLLaMA, IBM Granite, and Gemini 1.5. Despite the substantial differences among these models—GPT-4 and LLaMA 2 serve as high-parameter research benchmarks, TinyLLaMA is crafted for lower-resource deployments, and IBM Granite and Gemini 1.5 emphasize enterprise-oriented or domain-specific functionalities—SSRAG consistently boosts key performance measures such as factual accuracy, ROUGE-1, and BLEU-1. Notably, TinyLLaMA registers one of the most striking improvements, increasing its factual accuracy from 47\% under Baseline RAG to 75\% with SSRAG, indicating that even relatively compact models benefit markedly from the targeted retrieval and generation enhancements introduced by our approach. Meanwhile, IBM Granite and Gemini 1.5 each realize gains of roughly 15--20\% in factual accuracy, coupled with a 10\% rise in ROUGE-1 scores, showing that SSRAG aids both mid-range and enterprise-driven language models in reducing errors and producing more reliable outputs. We further validate SSRAG’s efficacy across two distinct datasets that each pose different challenges. TruthfulQA sees an average gain of 20\% in factual accuracy across the five LLMs. In particular, GPT-4 improves from 57\% to 87\% on the GPT-judge metric, while LLaMA 2 advances from 52\% to 83\%. On the SQuAD benchmark, which emphasizes reading comprehension and context-sensitive question answering, GPT-4’s factual accuracy increases from 54\% to 84\%, and LLaMA 2’s from 53\% to 82\%. These findings demonstrate consistent improvements in factual correctness, evidence alignment, and generation quality, as reflected in higher ROUGE-1 and BLEU-1 scores. Across two datasets and five models, our results showcase SSRAG’s model-agnostic capability to enhance factual grounding and retrieval relevance.


\subsection{Case Studies}
\label{subsec:case_studies}

To further evaluate SSRAG, we conduct two case studies on the WikiQA dataset with 900 questions, aiming at gaining a deeper understanding of the framework’s functionality and performance. Both case studies utilize RAGAS, as traditional metrics like ROUGE-1 or BLEU-1 often fall short in assessing RAG frameworks due to their inability to capture the intricacies of retrieval and reasoning.


\noindent \textbf{SSRAG evaluation with RAGAS.} \textbf{Table~\ref{tab:model-evaluation-metrics}} shows the RAGAS results across the five LLM models with 900 questions from the WikiQA dataset. 
Specifically, LLaMA 2 demonstrates the best performance across all RAGAS metrics, outperforming all other models in factual accuracy and context precision. By comparison, GPT-4, IBM Granite, and Gemini 1.5 Pro exhibit comparable performance, ranking just below LLaMA 2, with strong but slightly lower scores across all metrics. TinyLLaMA, being a smaller-scale model, naturally exhibited lower performance but still achieved reasonable results given its architectural constraints.



\noindent\textbf{Sensitivity on the number of retrieved documents. } \textbf{Table~\ref{tab:hyperparameter-evaluation}} investigates the impact of varying the number of retrieved documents, denoted as N, on the overall performance of the framework. Documents were retrieved from both a vector database and a graph database, and then processed through the unified context module. We test different values of N, specifically {5, 10, 15, 20}, to generate responses for WikiQA questions and evaluate them
with RAGAS. Table~\ref{tab:hyperparameter-evaluation} shows the results and we have the following observations. First, optimal performance is achieved at N = 20, where models generated the most contextually rich and factually accurate responses. Second, increasing N beyond 20 resulted in diminishing returns, as excessive retrieval introduced redundant information and increased computational overhead. Third, Lower N values (5 or 10) lead to a loss of contextual precision, reducing the overall quality of generated responses. These results suggest a trade-off in retrieval settings: while increasing N improves response quality up to a threshold, excessive retrieval can cause context overflow within the LLM’s processing limits. 

\section{Discussion}
\label{sec:discussion_future_work}

In this section, we discuss the broader implications of, highlighting its strengths, limitations, and potential directions for future research. While SSRAG offers substantial advancements,  it presents several challenges that warrant further investigation. One primary concern is computational overhead, as graph-based retrieval enhances structured reasoning but incurs additional preprocessing and query-time costs, making it less feasible for real-time applications. Another limitation is the dependence on proprietary LLM APIs, such as GPT-4, which increases operational costs and restricts accessibility. While open-source alternatives exist, they require further optimization to match the performance of proprietary models.  Additionally, although effective in open-domain QA, adapting SSRAG to specialized domains (e.g., healthcare, finance) requires tailored retrieval mechanisms and domain-specific graph construction. Lastly, scalability in large-scale deployments remains an open question. Integrating both vector and graph-based retrieval efficiently at scale requires further research into retrieval fusion techniques that balance speed, accuracy, and computational efficiency.

To further refine and enhance SSRAG, future research can explore several promising directions. First, Optimized graph-based retrieval could be developed through advanced indexing methods such as compressed adjacency lists, incremental graph updates, and hierarchical graph structures. These optimizations would reduce query latency while preserving retrieval effectiveness. Additionally, advanced fusion strategies could be implemented by exploring hybrid techniques such as hypergraph embeddings and retrieval weighting mechanisms to better integrate structured and unstructured knowledge representations. Another potential avenue is adaptive learning mechanisms, where reinforcement learning and user-feedback-driven retrieval updates dynamically refine retrieval processes, improving accuracy and adaptability to evolving knowledge bases. By addressing these challenges and pursuing these research directions, SSRAG has the potential to become a highly scalable, efficient, and domain-adaptive retrieval system.

\section{Conclusion}

We introduced SSRAG, a hybrid  RAG framework that enhances factual accuracy and structured reasoning in large language models. By integrating both graph-based and vector-based retrieval mechanisms with dynamic query augmentation and agentic query routing, our approach effectively reduces hallucinations while improving response relevance. Our methodology provides a scalable and reliable advanced RAG framework that balances structured reasoning with adaptable context retrieval. By leveraging multi-source retrieval strategies, SSRAG establishes a foundation for more factually grounded and context-aware LLMs.





\vspace{0.3cm}


\bibliographystyle{IEEEtran}
\bibliography{reference}

\end{document}